\pdfoutput=1
\documentclass[11pt]{article}

\usepackage[]{acl}

\usepackage{times}
\usepackage{latexsym}

\usepackage[T1]{fontenc}
\usepackage[utf8]{inputenc}

\usepackage{microtype}

\usepackage{algorithm}
\usepackage{algorithmic}

\usepackage{danudefs}
\usepackage{url}
\usepackage{hyperref}
\usepackage{multirow}
\usepackage{booktabs}
\usepackage{amsmath}
\usepackage{mathtools}
\usepackage[whole]{bxcjkjatype}

\title{Gender Bias in Masked Language Models for Multiple Languages}

\author{Masahiro Kaneko$^{1}$ \quad
        Aizhan Imankulova$^{2}$ \quad
        Danushka Bollegala$^{3,4}$\Thanks{ Danushka Bollegala holds concurrent appointments as a Professor at University of Liverpool and as an Amazon Scholar. This paper describes work performed at the University of Liverpool and is not associated with Amazon.} \quad
        Naoaki Okazaki$^{1}$ \\
        $^1$Tokyo Institute of Technology \quad
        $^2$CogSmart Co., Ltd. \\
        $^3$University of Liverpool \quad
        $^4$Amazon \\
        {\tt masahiro.kaneko@nlp.c.titech.ac.jp} \\
        {\tt aizhan.imankulova@cogsmart-global.com} \\
        {\tt danushka@liverpool.ac.uk} \quad
        {\tt okazaki@c.titech.ac.jp}
}

\begin{document}
\maketitle
\begin{abstract}
Masked Language Models (MLMs) pre-trained by predicting masked tokens on large corpora have been used successfully in natural language processing tasks for a variety of languages.
Unfortunately, it was reported that MLMs also learn discriminative biases regarding attributes such as gender and race.
Because most studies have focused on MLMs in English, the bias of MLMs in other languages has rarely been investigated.
Manual annotation of evaluation data for languages other than English has been challenging due to the cost and difficulty in recruiting annotators.
Moreover, the existing bias evaluation methods require the stereotypical sentence pairs consisting of the same context with attribute words (e.g. \textit{He/She is a nurse}).
We propose Multilingual Bias Evaluation (MBE) score, to evaluate bias in various languages using only English attribute word lists and parallel corpora between the target language and English without requiring manually annotated data.
We evaluated MLMs in eight languages using the MBE and confirmed that gender-related biases are encoded in MLMs for all those languages.
We manually created datasets for gender bias in Japanese and Russian to evaluate the validity of the MBE.
The results show that the bias scores reported by the MBE significantly correlates with that computed from the above manually created datasets and the existing English datasets for gender bias.
\end{abstract}

\section{Introduction}

Masked Language Models (MLMs)~\cite{devlin-etal-2019-bert}, which are pre-trained on large corpora, have been used successfully in natural language processing tasks for various languages~\cite{NEURIPS2019_c04c19c2,martin-etal-2020-camembert,conneau-etal-2020-unsupervised}.
Unfortunately, it has been reported that MLMs also learn social biases regarding attributes such as gender, religion, and race~\cite{kurita-etal-2019-measuring,dev2020measuring,kaneko-bollegala-2021-debiasing,10.1145/3442188.3445922}.
The bias in MLMs is evaluated by the imbalance of the likelihood between pairs of sentences associated with an attribute that has a common context (e.g. \textit{He/She is a nurse}).
\citet{nadeem-etal-2021-stereoset} masked the modified tokens (e.g. \textit{He, She}), and \citet{nangia-etal-2020-crows} masked the unmodified tokens (e.g. \textit{is, a, nurse}) one word at a time and calculated the likelihood from their predictions to evaluate the bias.
\citet{kaneko2021unmasking} evaluated the bias using the average of the likelihoods of all tokens without masking the MLM.

Despite the numerous studies of social bias in MLMs covering English, social biases in MLMs for other languages remain understudied~\cite{lewis2020gender,liang-etal-2020-monolingual,bartl-etal-2020-unmasking,zhao-etal-2020-gender}.
To realise the diverse and inclusive social and cultural impact of AI, we believe it is important to establish tools for detecting and mitigating unfair social biases in MLMs, not only for English but for all languages.
However, the significant manual annotation effort, the costs and difficulties in recruiting qualified annotators remain major challenges when creating bias evaluation benchmarks for target languages. 
For example, existing bias evaluation benchmarks such as CrowS-Pairs~\cite[\textbf{CP};][]{nangia-etal-2020-crows} and StereoSet~\cite[\textbf{SS};][]{nadeem-etal-2021-stereoset} require human-written sentences (or pairs of sentences) eliciting different types of social biases expressed in the target language.
However, scaling up this approach to all languages is challenging because recruiting a sufficiently large pool of annotators to cover the different types of social biases in those languages is difficult.
Because of the above-mentioned challenges, bias evaluation datasets and studies outside English remain under-developed.

\begin{figure*}[t]
  \centering
  \includegraphics[width=0.9\textwidth]{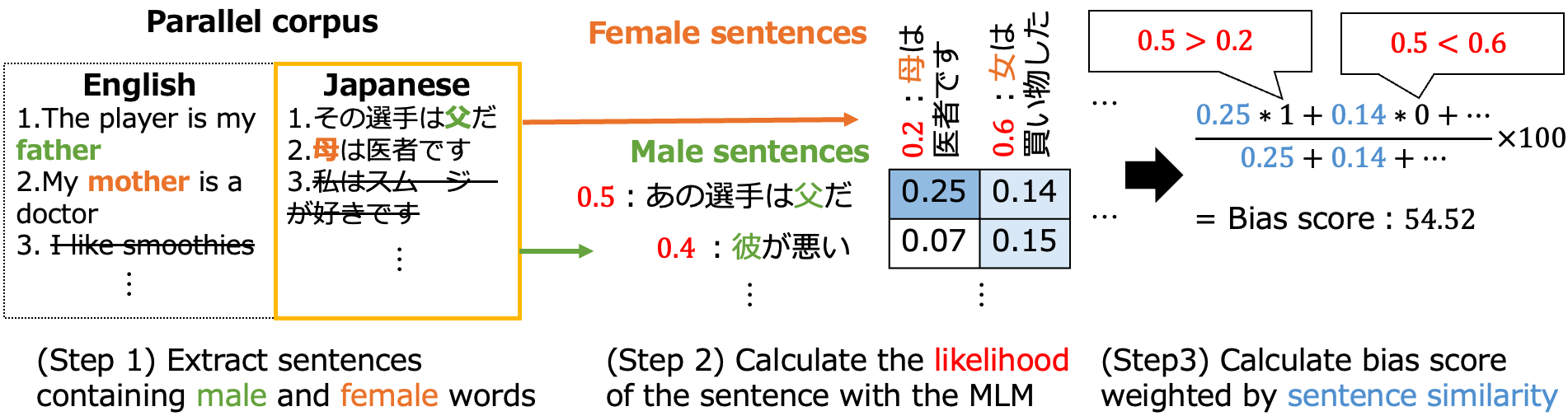}
  \caption{The bias evaluation method using a parallel corpus between English and the target language and an English female and male words list. Matrix values in Step 2 are the similarities between male and female sentences.}
  \label{fig:eval}
\end{figure*}

To address this problem, we propose Multilingual Bias Evaluation (MBE) score\footnote{Our code and dataset: \url{https://github.com/kanekomasahiro/bias_eval_in_multiple_mlm}}, a social bias evaluation method that can be used to evaluate biases in pre-trained MLMs for a target language without requiring significant manual annotations for that language. 
MBE can perform equivalent bias evaluation using only existing parallel corpora and lists of English masculine (e.g. \emph{he, his, father, son} etc.) and feminine (e.g. \emph{she, her, mother, daughter}, etc.) words, without requiring any manually annotated sentences for social biases in the target language.
Although MBE require parallel corpora, such sources already exist for numerous language pairs\footnote{\url{https://www.clarin.eu/resource-families/parallel-corpora}} or can be automatically mined with less effort compared to annotating bias evaluation data \cite{artetxe2019massively, artetxe-schwenk-2019-margin}.
As a concrete example, we evaluate the proposed method for gender bias, which exists in many languages.
Extending the proposed method to other types of social biases beyond gender biases is deferred to future work.
As shown in \autoref{fig:eval}, MBE first 
(shown in Step 1) extracts target language sentences containing female words (female sentences) and sentences containing male words (male sentences) from a parallel corpus between English and the target language using a list of gender words in English. 
Second, (shown in Step 2) MBE calculates the likelihoods for each of the extracted female and male sentences in the target language using the given MLM under evaluation. 
Finally, (shown in Step 3) MBE compares the likelihoods between each female sentence and all male sentences considering all pairwise combinations, and increment a count by 1 if the likelihood of the male sentence is greater than that of the female sentence, and 0 otherwise.

As for Step 1, we do not require any knowledge about the target language or manual annotations because we use only the existing English attribute word lists and parallel corpora between English and the target language.
This is attractive from a data availability point of view, which makes our proposed method easily extendable to different languages.
\citet{kaneko2021unmasking} found that the frequency of the words associated with the male gender to be significantly higher than that for the female gender in the data used to train MLMs.
They showed that these frequency-related biases are encoded into MLMs, and independently of whether a sentence contains a stereotypical or antistereotypical context, an MLM that is biased towards the male gender, on average, assigns higher likelihood scores to sentences that contain masculine words than feminine words\footnote{\citet{nangia-etal-2020-crows} define stereotypical sentence to be a case where an advantaged group (in the case of gender bias \emph{male} is considered as the advantaged group, whereas \emph{female} is the disadvantaged group) is associated with a pleasant attribute (e.g. \emph{The man is intelligent}) or a disadvantaged group is associated with an unpleasant attribute (e.g. \emph{The woman is careless}).}.
Inspired by this finding, in Step 2, we calculate the likelihood scores assigned by an MLM under evaluation to sentences that contain male and female related words in \emph{different} contexts  (e.g. \textit{He is a baseball player} and \textit{She is a nurse}). 
Step 3 of our proposed method performs the computation of the bias score considering the similarity between the contexts in sentence pairs that contain male and female related words.
The more similar the sentence pairs are, the more similar the estimates would be compared to the bias evaluation measures that require \emph{identical} contexts.
Therefore, we weight the bias score estimates by the similarity of the sentence pairs using the sentence representations obtained from the MLM under evaluation.
We ignore dissimilar sentence pairs and compute the bias score from the similar sentence pairs, which is defined as the percentage of sentence pairs where the sentence containing the masculine words is assigned a higher likelihood than the sentence containing the feminine words.

Bias in MLM is thought to depend on both the MLM and the evaluation data, so in this paper, we are investigating both using two corpora for English MLMs.
Using the proposed method, we evaluated gender bias in MLMs in eight other languages: German, Japanese, Arabic, Spanish, Portuguese, Russian, Indonesian, and Chinese.
Prior work investigating social biases in MLMs for English have shown that different types and levels of biases are shown by different MLMs even for the same language~\cite{kaneko-bollegala-2021-debiasing,kaneko2021unmasking,dev2020measuring}.
We defer covering different MLMs across multiple languages to future work and focus on establishing MBE as an evaluation measure that can be used for such a study.

Our evaluations show that all MLMs learn gender-related biases in all languages studied.
To further validate MBE, we conduct a meta-evaluation where we use an existing manually annotated English bias evaluation dataset and two additional datasets we annotate in this work covering gender-related biases in Japanese and Russian languages.
The bias scores computed using MBE show significantly high correlations with human bias annotations on both datasets (CP and SS), showing its validity for multiple languages as a gender bias evaluation method.
Furthermore, we show that MBE is superior to methods using machine translations to evaluate bias in non-English languages.
We also show that bias evaluation methods based on templates and word lists significantly overestimate the bias in MLMs due to the unnaturalness of the created templates.
Our analyses on the effects of English names on gender information and preservation of gender information in parallel corpora suggest that bias can be evaluated reasonably even with some loss of gender information.

\section{Related Work}

In the study of bias in English MLMs, \citet{may-etal-2019-measuring} and \citet{kurita-etal-2019-measuring} use a pair of artificial sentences created using manually written templates.
However, template-based evaluation is problematic because it uses an artificial context that does not reflect the natural usage and distribution of words in the target language.
To solve this problem, \citet{nadeem-etal-2021-stereoset} and \citet{nangia-etal-2020-crows} manually created bias evaluation datasets, SS and CP, respectively, with stereotypical and antistereotypical sentence-pairs with identical contexts, except the attribute words.
However, recent work has pointed out various issues in CP and SS datasets and has argued that they may not provide effective measurements of stereotyping~\cite{blodgett-etal-2021-stereotyping}.
In this study, (social) bias is defined as the tendency towards outputting sentences about a particular advantageous or disadvantageous group, such as males or females, given the same context by an MLM.
However, these benchmarks are currently the most commonly used benchmarks for bias evaluation in MLMs, so we also use them in this work. 
We note that MBE is \emph{independent} of any bias evaluation benchmark datasets.
Our focus in this paper is on evaluating gender bias in multiple languages and \emph{not} on comparing or proposing novel debiasing methods.
However, for the completion of the discussion, we note that methods for debiasing MLMs using sentence vectors from MLMs~\cite{bommasani-etal-2020-interpreting} and lists of English male and female words has been studied~\cite{sedoc-ungar-2019-role,kaneko-bollegala-2021-debiasing,dev2020measuring, Chou:ACL2022}.

In prior work on MLMs, social biases for languages other than English have rarely been investigated.
\citet{ahn-oh-2021-mitigating} investigated ethnic bias in monolingual MLM in six languages by extending the templates to other languages using machine translation.
The biases of MLMs have been evaluated using templates for English and Chinese \cite{liang-etal-2020-monolingual} and for English and German \cite{bartl-etal-2020-unmasking}.
\citet{zhao-etal-2020-gender} investigated the gender bias of a classifier that predicts the occupation from resumes using multilingual word embeddings and multilingual MLM embedding in Spanish, German and French.
They evaluated bias by using machine translation on the English data, when an MLM is used to create feature representations in a specific task.
However, this setting is different from that of our study, where we evaluate the bias of MLMs independently of a specific task.
Moreover, the above studies do not discuss or propose methods on how to create evaluation data that can be applied to many languages.

%coreference resolution\cite{rudinger-etal-2018-gender,zhao-etal-2018-gender}

Following the pioneering work by \citet{NIPS2016_a486cd07} that proposed a bias evaluation and debiasing methods, various studies have investigated social biases in English~\cite{caliskan2017semantics,zhao-etal-2018-learning,kaneko-bollegala-2019-gender,kaneko-bollegala-2021-dictionary,dev2019attenuating}. %wang-etal-2020-double,vargas-cotterell-2020-exploring,sweeney-najafian-2019-transparent,shin-etal-2020-neutralizing}.
Unlike the contextual word embeddings produced by MLMs, evaluating social biases in static word embeddings is relatively less complicated because it can often be done using word lists without requiring annotated sentences.
In static word embeddings, bias has been investigated in various languages besides English due to this ease of annotating evaluation data.
\citet{lauscher-glavas-2019-consistently} translated the English word lists into six languages and evaluated the bias of the word embeddings.
\citet{zhou-etal-2019-examining} proposed an evaluation metric for languages that require gender morphological agreement, such as in Spanish and French.
\citet{friedman-etal-2019-relating} quantified the gender bias of word embeddings to understand cultural contexts with large-scale data, and used it to characterize the statistical gender gap in education, politics, economics, and health in US states and several countries.
\citet{bansal2021debiasing} proposed a debiasing method by constructing the same bias space for multiple languages, and adapted it to three Indian languages.
Other bias studies have been conducted for specific languages~\cite{takeshita-etal-2020-existing,10.1145/3377713.3377792,sahlgren-olsson-2019-gender,chavez-mulsa-spanakis-2020-evaluating}, but they are not easily transferable to novel languages.

\section{Bias Evaluation for Multiple Languages}

Our proposed MBE score evaluates the gender bias of the target language under evaluation in three steps (see \autoref{fig:eval}). 
In Step 1, we first define the set of English sentences $\cE$ and the set of target language sentences $\cT$ of the parallel corpus, where $N$ is the data size, and $(e_i, t_i)$ is a parallel sentence pair.
Let $\cV_f$ (e.g. \textit{she, woman, female}) be the list of female words and $\cV_m$ (e.g. \textit{he, man, male}) be the list of male words in English.
We then extract sentences that contain a female or a male word from $\cE$.
Sentences that contain both male and female words are excluded.
Let us denote the set of sentences extracted for a female or a male word $w$ by $\Phi(w)$.
Let $\cE_f = \bigcup_{w \in \cV_f} \Phi(w)$ and $\cE_m = \bigcup_{w \in \cV_m} \Phi(w)$ be the sets of sentences containing respectively all of the male and female words.
The set of sentences in the target language of the source sentences included in $\cE_f$ and $\cE_m$ is denoted by $\cT_f$ and $\cT_m$, respectively.
It is assumed that gender information is retained in the parallel corpus, and whether this is actually the case is verified later.

In Step 2, we compute the likelihood for the full sentences in $\cT_f$ and $\cT_m$.
Let us consider a target sentence $T = w_1, w_2, \ldots, w_{|T|}$, containing length $|T|$ sequence of tokens $w_i$. 
We calculate the likelihood with All Unmasked Likelihood with Attention weights ~\cite[\textbf{AULA};][]{kaneko2021unmasking} which evaluates the bias by considering the weight of MLM attention as the importance of tokens.
Given an MLM with pre-trained parameters $\theta$, which we must evaluate it for its gender bias, let us denote the probability $P_{\mathrm{MLM}}(w_i | T; \theta)$ assigned by the MLM to a token $w_i$ conditioned on all the tokens of $T$. 
AULA predicts all of the tokens in $T$ using the attention weights to evaluate social biases considering the relative importance of words in a sentence, which is given by \eqref{eq:AULA}.
\begin{align}
    \label{eq:AULA}
    A(T) \coloneqq \frac{1}{|T|} \sum_{i=1}^{|T|} \alpha_i \log P_{\mathrm{MLM}}(w_i | T; \theta)
\end{align}
Here, $\alpha_i$ is the average of all multi-head attentions associated with $w_i$.

In Step 3, by comparing the likelihoods of female and male sentences returned by AULA, we calculate the bias score as the weighted average of the similarities of contexts using the sentence representations produced by the MLM under evaluation.
Specifically, We use the percentage of male ($T_m$) sentences (e.g. \textit{He is a baseball player}) preferred by the MLM over female ($T_f$) ones (e.g. \textit{She is a nurse}) to define the corresponding multilingual bias evaluation measure (\textbf{MBE bias score}) as follows:

{\small
\begin{align}
    \label{eq:score}
    100 \times \frac{\sum_{T_m \in \cT_m}\sum_{T_f \in \cT_f} C(T_m, T_f) \mathbb{I}(A(T_m) > A(T_f))}{\sum_{T_m \in \cT_m}\sum_{T_f \in \cT_f} C(T_m, T_f)}
\end{align}
}%
Here, $\mathbb{I}$ is the indicator function, which returns $1$ if its argument is True and $0$ otherwise.
$C(T_m, T_f)$ uses the average of the last layer in MLM for all tokens except special tokens to compute the sentence embeddings of $T_m$ and $T_f$ respectively and computes the cosine similarity of these embeddings.
According to this evaluation measure, values close to 50 indicate that the MLM under evaluation is neither females nor males biased, hence, it can be regarded as unbiased. 
On the other hand, values below 50 indicate a bias towards the male group and above 50 towards the female group.
We report a statistically significant difference comparing to the model with randomly assigned results of the indicator function $\mathbb{I}$ in \autoref{eq:score} with the McNemar's test ($p < 0.05$).
For each sentence, the presence or absence of bias is predicted by two methods, MLM and Random.
The McNemar’s test was used by classifying into four categories: only MLM was biased, only random was biased, both were unbiased, and both were biased.
We use the statistically significant difference to determine if there is a bias.

\section{Gender Bias in Masked Language Models}

\begin{table}[t]
%\begin{adjustbox}{width=\textwidth,center}
\centering
\small
\begin{tabular}{lcccccccc}
\toprule
Lang & TED & News \\
\midrule
%TED & 3K & 3K & 3K & 9K & 3K & 3K & 1K & 3K \\
German & 4.7K & 2.1K \\
Japanese & 6.2K & 1.8K \\
Arabic & 7.0K & 1.7K \\
Spanish & 7.1K & 17.3K \\
Portuguese & 5.7K & 2.2K \\
Russian & 6.7K & 3.9K \\
Indonesian & 2.9K & 0.5K \\
Chinese & 6.8K & 3.4K \\
\bottomrule
\end{tabular}
%\end{adjustbox}
\caption{The total number of male and female sentences extracted from the parallel data for each language.}
\label{tbl:data_size}
%\vspace{-3mm}
\end{table}

We use two parallel corpora, the TED2020 v1 corpus in the spoken language domain (\textbf{TED})\footnote{\url{https://opus.nlpl.eu/TED2020.php}} and the GlobalVoices corpus in the news domain (\textbf{News})\footnote{\url{https://opus.nlpl.eu/GlobalVoices-v2017q3.php}}.
\autoref{tbl:data_size} shows the total number of extracted male and female sentences for each language.
Except for Spanish, the News corpus is smaller than the TED corpus for all languages.
In particular, the Indonesian news corpus is an extremely low resource.
For the list of female and male words in English, we use the list created by \citet{NIPS2016_a486cd07}\footnote{\url{https://github.com/uclanlp/gn_glove/tree/master/wordlist}} in addition to the female and male names in CP~\cite{nangia-etal-2020-crows}.
The extracted male and female sentences were downsampled to create sets of an equal number of sentences.
We experimented on the GeForce RTX 2080 Ti using the \texttt{transformers}\footnote{\url{https://github.com/huggingface/transformers}} implementation with default settings~\cite{wolf-etal-2020-transformers}.
All evaluations are completed within 10 minutes.

We used Masked Language Models (\textbf{MLMs}) in eight languages for our experiments:
Japanese\footnote{\url{https://huggingface.co/cl-tohoku/bert-base-japanese-whole-word-masking}},
German\footnote{\url{https://huggingface.co/deepset/gbert-base}}~\cite{chan-etal-2020-germans},
Arabic\footnote{\url{https://huggingface.co/aubmindlab/bert-base-arabertv01}}~\cite{antoun-etal-2020-arabert},
Spanish\footnote{\url{https://huggingface.co/dccuchile/bert-base-spanish-wwm-uncased}}~\cite{CaneteCFP2020},
Portuguese\footnote{\url{https://huggingface.co/neuralmind/bert-base-portuguese-cased}}~\cite{souza2020bertimbau},
Russian\footnote{\url{https://huggingface.co/blinoff/roberta-base-russian-v0}}, 
Indonesian\footnote{\url{https://huggingface.co/cahya/bert-base-indonesian-522M}} and 
Chinese\footnote{\url{https://huggingface.co/hfl/chinese-bert-wwm-ext}}~\cite{cui-etal-2020-revisiting}.

% As English MLMs, we use BERT\footnote{\url{https://huggingface.co/bert-base-cased} and \url{https://huggingface.co/bert-large-uncased}}, multilingual BERT\footnote{\url{https://huggingface.co/bert-base-multilingual-uncased}}~\cite{devlin-etal-2019-bert}, RoBERTa\footnote{\url{https://huggingface.co/roberta-base} and \url{https://huggingface.co/roberta-large}}~\cite{liu2019roberta}, ALBERT\footnote{\url{https://huggingface.co/albert-base-v2}}~\cite{lan2019albert},
% DistilBERT\footnote{\url{https://huggingface.co/distilbert-base-cased}}, DistilRoBERTa\footnote{\url{https://huggingface.co/distilroberta-base}}~\cite{sanh2019distilbert}, ConvBERT\footnote{\url{https://huggingface.co/YituTech/conv-bert-medium-small}}~\cite{jiang2020convbert}, XLM\footnote{\url{https://huggingface.co/xlm-mlm-100-1280}}~\cite{NEURIPS2019_c04c19c2}, and Deberta\footnote{\url{https://huggingface.co/microsoft/deberta-xlarge-v2}}~\cite{he2020deberta}.

\begin{table}[t]
%\begin{adjustbox}{width=\textwidth,center}
\small
\centering
\begin{tabular}{lcc}
\toprule
Lang & MBE(TED) & MBE(News) \\
\midrule
German & 54.69$^\ddag$ & 55.12$^\ddag$ \\
Japanese & 54.52$^\ddag$ & 50.99 \\
Arabic & 55.72$^\ddag$ & 54.39$^\ddag$ \\
Spanish & 51.44$^\ddag$ & 51.69$^\ddag$ \\
Portuguese & 53.07$^\ddag$ & 54.99$^\ddag$ \\
Russian & 54.59$^\ddag$ & 51.00 \\
Indonesian & 52.38$^\ddag$ & 50.52 \\
Chinese & 52.86$^\ddag$ & 51.80$^\ddag$ \\
\bottomrule
\end{tabular}
%\end{adjustbox}
\caption{The bias score of MLMs using MBE in different languages. $\ddag$ indicates statistically significant difference at $p < 0.05$. }
\label{tbl:multi_bias}
%\vspace{-3mm}
\end{table}

\autoref{tbl:multi_bias} shows the bias scores of the proposed MBE method for the TED and News corpora for the MLMs considered.
Here, the significant difference is evaluated against the MBE score of a randomly assigned indicator function.
Overall, we see gender-related biases are reported in all cases.
In particular, significant biases are shown in the News corpus for all languages except Japanese, Russian and Indonesian.
%In particular, we can see that there is a bias in the male direction in TED for De, Japanese  Ar, and Russian.
Moreover, the different levels of biases reported for Russian and Japanese between TED and News corpora indicate that bias evaluations are affected not only by the MLMs but also the corpora used.
It is known that the bias tendency of MLMs changes depending on the training data~\cite{10.1145/3366424.3383559}, and similarly, the bias evaluation of MLMs is affected by the evaluation corpus.
Because MBE can evaluate bias in various domains as long as there are parallel corpora. It can also capture corpus-dependent biases, unlike existing methods requiring manually created domain-specific sentence pairs.

\section{Meta-Evaluation}

We perform a meta-evaluation to validate MBE scores against human bias ratings.
In \S\ref{sec:manual-english} we measure the correlation between MBE scores and existing measures on CP and SS, which are manually annotated bias evaluation benchmarks for English.
In \S\ref{sec:jp-ru} to compare the evaluation methods in the target languages using MBE and manually annotated data, we manually translate the CP into the Japanese and Russian, which demonstrate high corpus-specific biases according to \autoref{tbl:multi_bias}.
%In English, we compare the likelihood of male and female sentences and investigate the validity of using sentence vectors in the proposed method.
%We also show that the proposed method outperforms the machine translation and template methods in Japanese and Russian.

\subsection{Gender Bias Evaluation Using Manually Annotated Data in English}
\label{sec:manual-english}

\begin{table}[t]
\small
\centering
\begin{tabular}{clcc}
\toprule
& & Shf & MBE \\
\midrule
\multirow{4}{*}{CP}
& \textit{Spearman} & 0.06 & \textbf{0.41} \\
& \textit{Pearson}  & 0.05 & \textbf{0.63}$^\dagger$ \\
& \textit{Direction} & 0.54 & \textbf{0.72} \\
& \textit{Diff} & 4.06 & \textbf{2.36} \\
\midrule
\multirow{4}{*}{SS}
& \textit{Spearman} & 0.21 & \textbf{0.41}  \\
& \textit{Pearson} & 0.04 & \textbf{0.62}$^\dagger$  \\
& \textit{Direction} & 0.54 & \textbf{0.72} \\
& \textit{Diff} & 6.66 & \textbf{5.04} \\
\bottomrule
\end{tabular}
%\end{adjustbox}
\caption{Bias scores computed using Shf and the MBE methods for English MLMs in CP and SS. Correlation between the original and proposed evaluation represented by \textit{Spearman} and \textit{Pearson} correlation coefficients. $\dagger$ indicates significant correlation at $p < 0.05$. \textit{Direction} is the percentage of agreement for direction of the bias score between original and proposed evaluations. \textit{Diff} is the mean of the difference between the bias scores of the original and proposed methods. }
\label{tbl:cpss_bias}
%\vspace{-3mm}
\end{table}

To validate MBE scores using human bias ratings, we use CP and SS datasets for \textit{English}.
As baseline method we use 
%(a) Unweighted Averaging (\textbf{Avg}), which does not use $C(T_m,T_f)$ in \eqref{eq:score},
%(b) \textbf{Edit}, which calculates similarity $C(T_m,T_f)$ using the edit distance of sentence pairs, and
\textbf{Shf}, which shuffles the sets of male and female sentences and randomly pair sentences from this set.
Shf is used to show the usefulness of comparing the likelihoods of male and female sets.
%Avg is used to test the effectiveness of prioritizing the scores of similar sentences.
%Unlike MBE, Edit does not use sentence embeddings extracted from MLMs in the bias score computation.
%Therefore, by comparing MBE against Edit, we can evaluate any effect due to biases in the MLM when computing bias scores.
%Edit is used to show that any biases already encoded in the sentence embeddings created from MLM is not noise in the evaluation.
In the existing evaluation method using manually annotated sentence pairs, the bias score is calculated for stereotypical $S_s$ (e.g. \textit{He is a doctor}) and anti-stereotypical $S_a$ (e.g. \textit{She is a doctor}) sentences with identical contexts as follows:
\begin{align}
\label{eq:AUL}
    \frac{100}{N} \sum_{S_s, S_a} \mathbb{I}(A(S_s) > A(S_a))
\end{align}
where $N$ is the total number of sentences. 
We use this bias score as an upper bound score to compare against it the results for Shf and MBE using the rank correlations (\textit{Spearman} and \textit{Pearson}), the agreement of the direction of bias between female and male directions (\textit{Direction}), where the bias scores above 50 indicate a bias towards the male direction and that below 50 towards the female direction, and the difference of the bias scores (\textit{Diff}) from the results of the method using manual annotation.
In the proposed method and Shf, for the gender bias data of CP and SS, we extract sentences containing male and female words for each sentence, instead of sentence pairs, and use them for evaluation using \autoref{eq:AUL}.

% As English MLMs\footnote{Links for the models below are listed in Supplementary.}, we use BERT, multilingual BERT~\cite{devlin-etal-2019-bert}, RoBERTa~\cite{liu2019roberta}, ALBERT~\cite{lan2019albert},
% DistilBERT, DistilRoBERTa~\cite{sanh2019distilbert}, ConvBERT~\cite{jiang2020convbert}, XLM~\cite{NEURIPS2019_c04c19c2}, and Deberta~\cite{he2020deberta}.
As English MLMs, we use BERT\footnote{\url{https://huggingface.co/bert-base-cased} and \url{https://huggingface.co/bert-large-uncased}}, multilingual BERT\footnote{\url{https://huggingface.co/bert-base-multilingual-uncased}}~\cite{devlin-etal-2019-bert}, RoBERTa\footnote{\url{https://huggingface.co/roberta-base} and \url{https://huggingface.co/roberta-large}}~\cite{liu2019roberta}, ALBERT\footnote{\url{https://huggingface.co/albert-base-v2}}~\cite{lan2019albert},
DistilBERT\footnote{\url{https://huggingface.co/distilbert-base-cased}}, DistilRoBERTa\footnote{\url{https://huggingface.co/distilroberta-base}}~\cite{sanh2019distilbert}, ConvBERT\footnote{\url{https://huggingface.co/YituTech/conv-bert-medium-small}}~\cite{jiang2020convbert}, XLM\footnote{\url{https://huggingface.co/xlm-mlm-100-1280}}~\cite{NEURIPS2019_c04c19c2}, and Deberta\footnote{\url{https://huggingface.co/microsoft/deberta-xlarge-v2}}~\cite{he2020deberta}.
Since BERT and RoBERTa each use two models of different sizes, we use a total of 11 models.
We report the averaged results over the above 11 models.

%\autoref{tbl:cpss_bias} shows the comparison results with the method using manually annotated data.
\autoref{tbl:cpss_bias} shows that MBE has high performance in all evaluations.
%Shf has the worst performance in all evaluations, indicating that comparing male and female sentences separately leads to a more effective bias evaluation.
Performance of Shf highlights the importance of comparing male against female sentences in sentence pairs.
%Compared to Avg, MBE performed better, especially in \textit{Direction}, while Edit had the same number of highest scores as Avg, indicating that it did not improve much in terms of performance.
%This indicates that it is effective to use sentence vectors to decompose the scores of similar sentences.
%MBE consistently outperforms \textbf{Avg}, indicating the importance of considering the similarity between contexts.
%Moreover, the better performance of MBE over \textbf{Edit} shows that using sentence embeddings from MLMs provides more reliable estimates of contextual similarity.

\begin{table}[t]
%\begin{adjustbox}{width=\textwidth,center}
\centering
\small
\begin{tabular}{llcc}
\toprule
& MLM & Bias score & Diff \\
\midrule
\multirow{4}{*}{HT(Japanese)}
& base-subword & 52.67$^\ddag$ & - \\
& large-subword & 56.87$^\ddag$ & - \\
& base-char & 48.47$^\ddag$ & - \\
& large-char & 55.73$^\ddag$ & - \\
\midrule
\multirow{4}{*}{MT(Japanese)}
& base-subword & 49.24 & -3.43 \\
& large-subword & 52.67$^\ddag$ & -4.20 \\
& base-char & 54.20$^\ddag$ & 5.73 \\
& large-char & 45.80$^\ddag$ & 9.93 \\
\midrule
\multirow{4}{*}{MBE(Japanese)}
& base-subword & 54.89$^\ddag$ & \textbf{2.22} \\
& large-subword & 55.85$^\ddag$ & \textbf{-1.02} \\
& base-char & 52.69$^\ddag$ & \textbf{4.22} \\
& large-char & 50.60 & \textbf{-5.13} \\
\midrule
\multirow{4}{*}{Tmp(Japanese)}
& base-subword & 88.31$^\ddag$ & 35.64 \\
& large-subword & 82.13$^\ddag$ & 25.26 \\
& base-char & 64.63$^\ddag$ & 16.16 \\
& large-char & 45.40$^\ddag$ & -10.33 \\
\bottomrule
\end{tabular}
%\end{adjustbox}
\caption{The CP bias scores for manually translated CP to Japanese  and bias scores for machine translated CP and the proposed method MBE. Diff shows the difference between MT, MBE and Tmp bias scores and HT bias scores, respectively. $\ddag$ indicates statistically significant difference at $p < 0.05$.}
\label{tbl:ja_eval}
%\vspace{-3mm}
\end{table}

\begin{table}[t]
%\begin{adjustbox}{width=\textwidth,center}
\centering
\small
\begin{tabular}{llcc}
\toprule
& MLM & Bias score & Diff \\
\midrule
\multirow{2}{*}{HT(Russian)}
& wiki\&nwes & 46.95$^\ddag$ & - \\
& subtitle\&sns & 48.85$^\ddag$ & - \\
\midrule
\multirow{2}{*}{MT(Russian)}
& wiki\&nwes & 49.62 & 2.67 \\
& subtitle\&sns & 50.38 & 1.53 \\
\midrule
\multirow{2}{*}{MBE(Russian)}
& wiki\&nwes & 46.05$^\ddag$ & \textbf{-0.90} \\
& subtitle\&sns & 48.82$^\ddag$ & \textbf{-0.03} \\
\midrule
\multirow{2}{*}{Tmp(Russian)}
& wiki\&nwes & 34.87$^\ddag$ & -12.1 \\
& subtitle\&sns & 63.51$^\ddag$ & 14.7 \\
\bottomrule
\end{tabular}
%\end{adjustbox}
\caption{The CP bias scores for manually translated CP to Russian and bias scores for machine translated CP and the proposed method MBE. Diff shows the difference between MT, MBE and Tmp bias scores and HT bias scores, respectively. $\ddag$ indicates statistically significant difference at $p < 0.05$.}
\label{tbl:ru_eval}
%\vspace{-3mm}
\end{table}

\subsection{Gender Bias Evaluation Using Manually Annotated Data in Japanese and Russian}
\label{sec:jp-ru}

To validate MBE scores, which does not require evaluation data with identical context, nor manual creation of evaluation data in the target languages \textit{other than English}, we use the following methods:

\noindent
\textbf{HT:} Native speakers manually translated all 262 sentence pairs in CP into Japanese and Russian and apply \autoref{eq:AUL}.
This human translated (\textbf{HT}) baseline can be seen as an upper bound for bias evaluation compared to MBE, which does not require translated examples.
Lower difference from these bias scores in this human-translated (HT) method would indicate a more reliable bias evaluation measure.
Note that, it is not appropriate to compare the bias score calculated using the English MLMs with the bias score calculated using the Japanese MLMs because we are evaluating different models. Therefore, we calculate the bias score in \autoref{eq:AUL} using the data translated into Japanese and Russian.
%First, let \textbf{HT} be a method that manually translates CP data into the target language and evaluates MLMs using those sentence pairs with identical context.

\noindent
\textbf{MBE:} Here, we let \textbf{MBE(Japanese)} and \textbf{MBE(Russian)} be the MBE scores computed using the \autoref{eq:score} and parallel data created above by manually translating original (English) CP dataset into Japanese and Russian for Step 1, respectively.
%\textbf{MBE(Japanese)} and \textbf{MBE(Russian)} use the English gender word list, manually translated respectively into Japanese and Russian. 
%We also manually translated the English gender word list into Japanese and Russian, which were used for \textbf{MBE(Japanese)} and \textbf{MBE(Russian)}.

\noindent
\textbf{MT:} As an alternative to costly manual translations, we use Google Machine Translation method (\textbf{MT})\footnote{In July 2021, we translated CP data using google spreadsheet function: \url{https://support.google.com/docs/answer/3093331?hl}} to translate sentence pairs in CP sharing identical contexts into each target language and apply \autoref{eq:AUL}.
%So we can use sentence pairs with identical contexts that have been machine translated into the target language.

\noindent
\textbf{Tmp:} Although it requires some knowledge about the target language, one can create templates in the target language for both genders such as ``[Gender]は[Occupation]です'' (\textit{[Gender] is a/an [Occupation]}) in Japanese, and fill in male and female word pairs, and occupation words as in ``彼/彼女は医者です'' (\textit{He/She is a doctor}) to create an equal number of sentences as the evaluation data for \autoref{eq:AUL}.
In the template-based method (\textbf{Tmp}), five word pairs were used for Japanese and Russian following prior work by \citet{kurita-etal-2019-measuring}\footnote{Japanese:
彼:彼女, 男:女, 父:母, 兄:姉, 叔父:叔母. Russian: Он:Она, Мужчина:Женщина, Папа:Мама, Брат:Сестра, Дядя:Тетя (English: He:She, Man:Woman, Father:Mather, Brother:Sister, Uncle:Aunt)}.
The templates were ``[Gender]は[Occupation]です。'' and ``[Gender]は[Occupation]に興味がある。'' in Japanese and ``[Gender] - [Occupation].'' and ``[Gender] - [Occupation] по специальности.'' were used for Russian.
We extracted respectively 644 and 154 occupation words for Japanese and Russian from Wikipedia\footnote{\url{https://ja.wikipedia.org/wiki/}職業一覧 and \url{https://ru.wikipedia.org/wiki/}Категория:Профессии}.
Following prior work by \citet{kurita-etal-2019-measuring}, we generated respectively 6400 and 1500 template sentences for Japanese and Russian, and evaluated them using sentence pairs with identical contexts.
%These templates are based on the study of English templates~\cite{kurita-etal-2019-measuring}.

For Japanese MLMs, we evaluate four Japanese BERT models (\textbf{base-subword}\footnote{\url{https://huggingface.co/cl-tohoku/bert-base-japanese-v2}}, \textbf{large-subword}\footnote{\url{https://huggingface.co/cl-tohoku/bert-large-japanese}}, \textbf{base-char}\footnote{\url{https://huggingface.co/cl-tohoku/bert-base-japanese-char-v2}}, \textbf{large-char}\footnote{\url{https://huggingface.co/cl-tohoku/bert-large-japanese-char}}), subword-based and character-based, with base and large sizes.
For Russian, we use two MLMs -- one trained on Wikipedia and news data (\textbf{wiki\&news})\footnote{\url{https://huggingface.co/DeepPavlov/rubert-base-cased}} and the other on OpenSubtitles~\cite{lison-tiedemann-2016-opensubtitles2016} and SNS data~\cite{shavrina2017methodology}.
% For Japanese MLMs, we evaluate four Japanese BERT models\footnote{Links for the models below are listed in appendix.} (\textbf{base-subword}, \textbf{large-subword}, \textbf{base-char}, \textbf{large-char}, subword-based and character-based, with base and large sizes.
% For Russian, we use two MLMs -- one trained on Wikipedia and news data (\textbf{wiki\&news}) and the other on OpenSubtitles~\cite{lison-tiedemann-2016-opensubtitles2016} and SNS data~\cite{shavrina2017methodology}.
% For Japanese MLMs, we evaluate four Japanese BERT models (\textbf{base-subword}\footnote{\url{https://huggingface.co/cl-tohoku/bert-base-japanese-v2}}, \textbf{large-subword}\footnote{\url{https://huggingface.co/cl-tohoku/bert-large-japanese}}, \textbf{base-char}\footnote{\url{https://huggingface.co/cl-tohoku/bert-base-japanese-char-v2}}, \textbf{large-char}\footnote{\url{https://huggingface.co/cl-tohoku/bert-large-japanese-char}}), subword-based and character-based, with base and large sizes.
% For Russian, we use two MLMs -- one trained on Wikipedia and news data (\textbf{wiki\&news})\footnote{\url{https://huggingface.co/DeepPavlov/rubert-base-cased}} and the other on OpenSubtitles~\cite{lison-tiedemann-2016-opensubtitles2016} and SNS data~\cite{shavrina2017methodology}(\textbf{subtitle\&sns})\footnote{\url{https://huggingface.co/DeepPavlov/rubert-base-cased-conversational}}.
For Japanese and Russian, we use the difference of the bias scores instead of the correlation coefficients with HT because the number of publicly available pre-trained MLMs is smaller than that of English.

Tables~\ref{tbl:ja_eval} and \ref{tbl:ru_eval} show the bias scores of HT, MT, Tmp and MBE and their differences measured against HT for Japanese and Russian MLMs, respectively.
We see that the difference between the bias scores of HT and MBE are smaller than that for MT, indicating that MBE closely approximates the human bias ratings in HT than other alternatives.
Moreover, we see that the direction of bias is reversed for base-char, large-char, and subtitle\&sns compared to HT.
Note that we can not directly compare Tmp with other methods due to the difference in evaluation data.
However, as one of the previous bias evaluation methods, Tmp overestimates the biases of MLMs, especially for Japanese subwords.
This is because simple artificial templates often over-emphasize gender biases compared to natural sentences, 
%This may be because the effect of stereotyping is emphasized more than in actual sentences since the evaluation is done with simple, artificial sentences.
Interestingly, MBE is more accurate than MT when evaluating gender biases.
Further investigations revealed that MT model itself could produce gender-biased translations, thereby adding noise to the translated sentences. 
%Moreover, MBE can evaluate the bias more accurately than MT.
%This may be due to the influence of gender bias in the MT model itself and translation failures, which contribute to the noise in the evaluation.

%\section{Language-Specific Aspects of Bias}
%\label{sec:analysis}

%In this section, we investigate the identification of gender using names of people and whether gender information in English is retained in the target language of the parallel corpora used for Step 1.

\section{Bias in Personal Names}

\begin{table}[t]
%\begin{adjustbox}{width=\textwidth,center}
\centering
\small
\begin{tabular}{llcc}
\toprule
& MLM & Bias score & Diff \\
\midrule
\multirow{4}{*}{HT$_{\rm name}$(Japanese)}
& base-subword & 52.29$^\ddag$ & -0.38 \\
& large-subword & 54.58$^\ddag$ & -2.29 \\
& base-char & 48.47$^\ddag$ & 0.00 \\
& large-char & 53.44$^\ddag$ & -2.29 \\
\midrule
\multirow{2}{*}{HT$_{\rm name}$(Russian)}
& wiki\&news & 47.33$^\ddag$ & 0.38 \\
& subtitle\&sns & 48.09$^\ddag$ & -0.76 \\
\bottomrule
\end{tabular}
%\end{adjustbox}
\caption{The difference between the bias score for the original data and the bias score for the CP data translated into Japanese and Russian with the names of people replaced by Japanese and Russian, respectively. $\ddag$ indicates statistically significant difference at $p < 0.05$.}
\label{tbl:name_eval}
%\vspace{-3mm}
\end{table}

One of the most significant differences in the frequency of words used in each language that affects gender bias is the names of people.
In bias evaluation, male and female names are used to identify the gender~\cite{caliskan2017semantics,romanov-etal-2019-whats}.
However, when names are transliterated from English to the target language, those transliterated names might be infrequent in the target language and might not be gender representative.
To study the effect of this issue on gender bias evaluation, we conduct the following experiment.
First, for the Japanese and Russian target languages, we replace the transliterated English names in the CP data with native Japanese and Russian names of the same gender.
Next, we compare the bias scores with those before the replacement in Tables~\ref{tbl:ja_eval} and \ref{tbl:ru_eval}.
We extracted the top 10 most popular names among Japanese\footnote{\url{https://www3.nhk.or.jp/news/special/sakusakukeizai/articles/20181127.html}} and Russians\footnote{\url{https://znachenie-tajna-imeni.ru/top-100-zhenskih-imen/} and \url{https://znachenie-tajna-imeni.ru/top-100-muzhskih-imen/}} for both genders, and randomly substituted them with the transliterated English names.
For example, we rewrite  ``\underline{シェリー}はナースです'' $\rightarrow$ ``\underline{美咲}はナースです'' (``\underline{Shelly} is a nurse'' $\rightarrow$ ``\underline{Misaki} is a nurse'').

\autoref{tbl:name_eval} shows the MBE score for Japanese (HT$_{\rm name}$(Japanese)), and Russian (HT$_{\rm name}$(Russian)) after the name replacement and the corresponding differences w.r.t. original bias scores shown in Tables~\ref{tbl:ja_eval} and \ref{tbl:ru_eval}).
%\autoref{tbl:name_eval} shows the difference between the result of replacing the English names with native Japanese and Russian names, and the results HT(Japanese) and HT(Russian) before the replacement in \autoref{tbl:ja_eval} and \autoref{tbl:ru_eval}, respectively.
We can see that the bias scores of the Japanese base models and all the Russian models are almost the same compared to respective values in Tables \ref{tbl:ja_eval} and \ref{tbl:ru_eval}.
The large models for Japanese differ by about -2.29, which is lower than the baseline in the table.
Moreover, the direction of the bias has not changed in both languages compared to respective directions in Tables \ref{tbl:ja_eval} and \ref{tbl:ru_eval}.
These results suggest that the bias can be evaluated reasonably even when English names are transliterated into a target language.

\section{Preserving Gender in Parallel Corpora}

\begin{figure}[t]
  \centering
  \includegraphics[width=0.95\columnwidth]{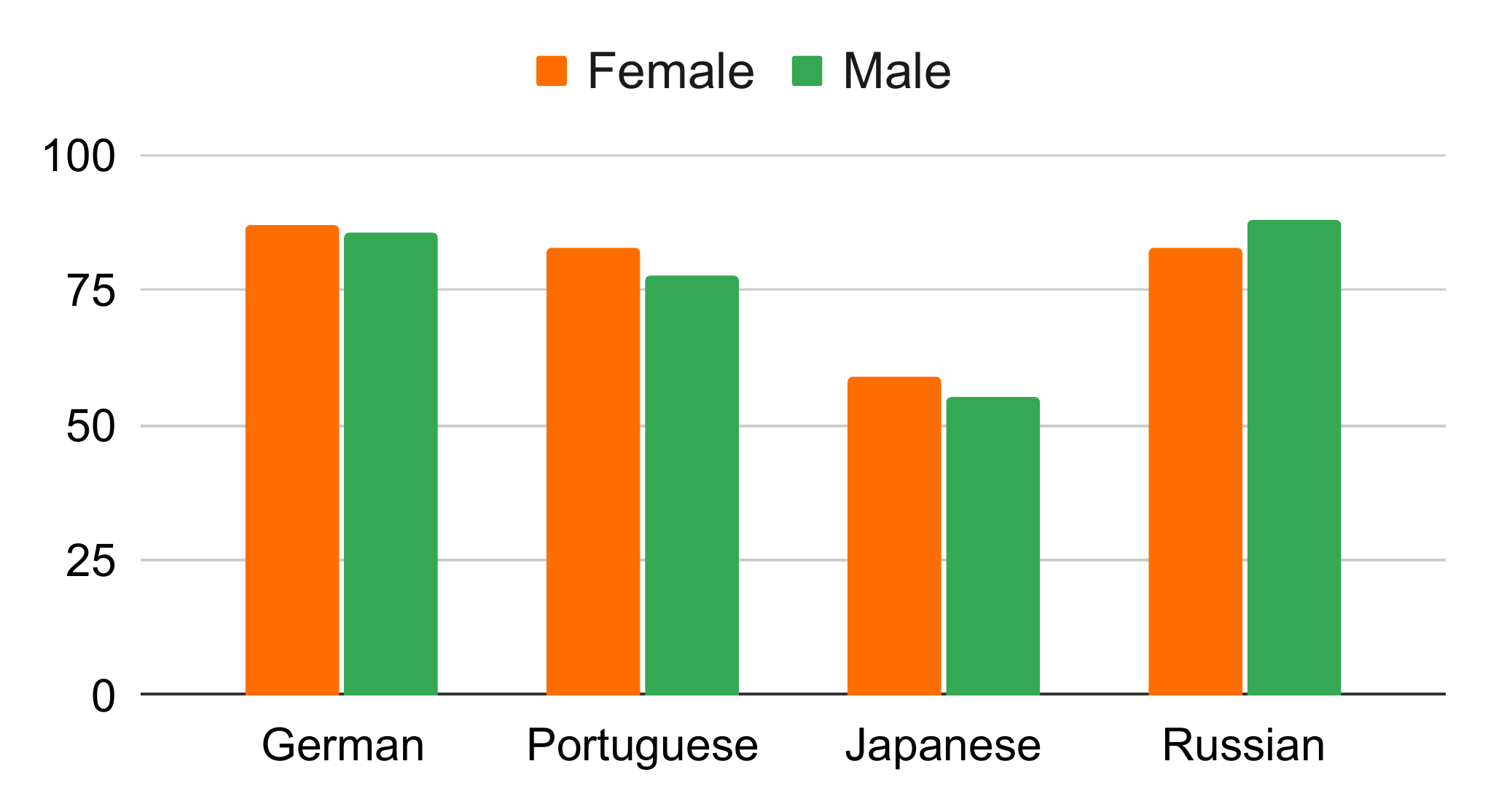}
  \caption{Percentage of manually translated sentences preserving gender information from English News data.}
  \label{fig:gender}
\end{figure}

Step 1 of the proposed method requires that gender information in English (source) sentence matches that with the target translation in the parallel data.
To test for this, we examine the proportion of sentences in which the corresponding translated words of English ``she'' and ``he'' appear to determine whether female or male gender information is retained.
%To investigate whether gender information is retained in the parallel corpora used for the Step 1 of the proposed method, we examine the proportion of sentences in which the corresponding translated words of English ``she'' and ``he'' appear to determine whether female or male gender information is retained.
We use the News corpus and select Japanese and Russian, which had no bias, and German and Portuguese, which had significant biases (\autoref{tbl:multi_bias}).

\autoref{fig:gender} shows the percentage of sentences where gender was retained for male and female sentences in the target languages\footnote{This is a conservative underestimate of gender preservation, because gender words can be translated by paraphrasing.}.
For German, Portuguese, and Russian, gender is retained in more than 80\% of the sentences.
This suggests that when the the percentage of gender-preserved sentences is large, it does not affect the MBE score.
In Japanese, gender information is retained in only about 60\% of sentences, which is much lower than in other languages.
This may be because Japanese is a null-subject language that allows independent clauses to omit explicit subjects.
In fact, in some cases, gender words were omitted in the parallel corpus, for example ``He owns a grocery store and runs a motorcycle rental business.'' was translated to ``自分の食料品店を持ち、レンタルバイクビジネスも営んでいる。 (Owns a grocery store and runs a rental motorcycle business.)''.
Contrarily, from the results in \autoref{tbl:ja_eval}, MBE(Japanese) can detect the bias better than other methods.
The reason may be that even if the gender words are omitted if the context is composed of words that often co-occur with male and female words, it is possible that it complements the gender information.
In fact, \citet{NIPS2016_a486cd07} show that words that co-occur with male and female words retain gender information.
The results also show that gender preservation is not heavily biased in either the male or female direction, based on the small difference between percentages for male and female sentences for each language.
This suggests that the bias in the preservation of gender information may not affect the evaluation of the proposed method.

\section{Conclusion}
\label{seq:coclusion}

% In order to evaluate discriminatory bias of MLMs in languages other than English, we proposed a method to construct bias evaluation data for the target language as long as there is a parallel corpus of English and the target language and a list of female and male words in English.
In this paper, we showed that a bias evaluation data and evaluation of MLMs for discriminatory bias can be systematically created as long as there is a parallel corpus of English and the target language and a list of female and male words in English.
Our meta-evaluation proved that the proposed multilingual bias evaluation method could perform correct evaluation comparing against method using manually created data, at least for Russian, Japanese, and English.
The experimental results show that gender bias exists in all eight languages of our experiments.
We also showed that the proposed method is superior to the methods that use machine translation to translate the English bias evaluation data into the target language and the methods that use templates and word lists.

\section*{Acknowledgements}

This paper is based on results obtained from a project, JPNP18002, commissioned by the New Energy and Industrial Technology Development Organization (NEDO).

\bibliography{arr}
\bibliographystyle{acl_natbib}

\end{document}